\documentclass[10pt, journal]{IEEEtran}
\ifCLASSINFOpdf
\else
\fi
\usepackage{graphicx}
\usepackage{mathbbol}
\usepackage[lined,ruled,vlined]{algorithm2e}
\usepackage{xcolor}
\usepackage{multirow}
\usepackage{amsmath}
\usepackage{color,soul}
\usepackage{geometry}
\usepackage{amssymb}
\usepackage[numbers]{natbib}
\usepackage{lineno}
\usepackage{colortbl}
\usepackage{multirow}
\usepackage{amsmath}
\usepackage{float}
\usepackage{textcomp}
\usepackage{titlesec}
\usepackage[ruled,vlined]{algorithm2e}
\usepackage{multirow}
\usepackage[nolist,nohyperlinks]{acronym}
\usepackage{adjustbox}

\begin{document}
	\title{CRL+: A Novel Semi-Supervised Deep Active Contrastive Representation Learning-Based  Text Classification Model for Insurance Data}
	 \author{Amir~Namavar Jahromi,~\IEEEmembership{Student Member, ~IEEE},
        Ebrahim~Pourjafari,
        Hadis~Karimipour,~\IEEEmembership{Senior Member,~IEEE,}
		Amit~Satpathy,
		and~Lovell~Hodge
 	\thanks{Amir~Namavar Jahromi and Hadis Karimipour are with the Schulich School of Engineering, University of Calgary, Alberta, Canada (email: amir.namavarjahromi@ucalgary.ca and hadis.karimipour@ucalgary.ca).}
 	\thanks{Ebrahim Pourjafari, Amit Satpathy, and Lovell Hodge are with the Munich Re Canada (email: EPourjafari@munichre.ca, ASatpathy@munichre.ca, and LHodge@munichre.ca)}
	 }
\maketitle
	
\begin{abstract}
    \textcolor{black}{Financial sector and especially the insurance industry} collect vast volumes of text on a daily basis and through multiple channels (their agents, customer care centers, emails, social networks, and web in general). The information collected includes policies, expert and health reports, claims and complaints, results of surveys, and relevant social media posts. It is difficult to effectively extract label, classify, and interpret the essential information from such varied and unstructured material. Therefore, the Insurance Industry is among the ones that can benefit from applying technologies for the intelligent analysis of free text through \textcolor{black}{Natural Language Processing (NLP)}.

    In this paper, CRL+, a novel text classification model combining Contrastive Representation Learning (CRL) and \textcolor{black}{Active Learning} is proposed to handle the challenge of using semi-supervised learning for text classification. In this method, supervised (CRL) is used to train a RoBERTa transformer model to encode the textual data into a contrastive representation space and then classify using a classification layer. This (CRL)-based transformer model is used as the base model \textcolor{black}{in the proposed} Active Learning mechanism to classify all the data in an iterative manner.
    The proposed model is evaluated using unstructured obituary data with objective to determine the cause of the death from the data. This model is compared with the CRL model and an Active Learning model with the RoBERTa base model. The experiment shows that the proposed method can outperform both methods for this specific task. 
\end{abstract}
	\IEEEpeerreviewmaketitle
	
    \begin{acronym}

     \acro{ML}{Machine Learning}
     \acro{NLP}{Natural Language Processing}
     \acro{PLM}{Pre-trained Language Model}
     \acro{RNN}{Recurrent Neural Network}
     \acro{CRL}{Contrastive Representation Learning}
     \acro{TP}{True Positive}
     \acro{TN}{True Negative}
     \acro{FP}{False Positive}
     \acro{FN}{False Negative}
     \acro{MCC}{Matthews Correlation Coefficient}
     \acro{LSTM}{Long Short-Term Memory} 
     \acro{CNN}{Convolutional Neural Network}
     \acro{NB}{Na\"ive Bayes}
     \acro{SVM}{Support Vector Machine}
     \acro{DNN}{Deep Neural Network}
     \acro{QA}{Question Answering}
     \acro{VDCNN}{Very Deep Convolutional Neural Network}
     \acro{MLM}{Masked Language Modeling}
     \acro{NLI}{Natural Language Inference}

    \end{acronym}
    
\section{Introduction}
Text classification is a classical problem in \ac{NLP} that aims to assign a label to textual units like words, sentences, paragraphs, or documents \cite{p1}. 
Text classification has a wide range of applications. It is used in question answering, spam detection, sentiment analysis, news categorization, user intent detection, and many others. \ac{NLP} applications are becoming more popular daily due to 
 the advances in various computational linguistics and the abundance of training data from websites, personal communications (emails, text messages), social media, tickets, insurance claims, user reviews, and questions/answers from customer services. 
The insurance industry collects a large amount of textual data as part of their day to day processes. The vast majority of this textual data is read and interpreted by human agents resulting in higher costs and slower processes. This creates an opportunity to automate some text processing tasks and make the insurance processes faster and cheaper. 
Traditional \ac{NLP} techniques will assist insurance companies to personalize insurance products and better respond to existing and future clients’ needs. More accurate underwriting models and constant learning from new data will be reflected in premiums that are often conservatively overestimated \textcolor{black}{due to} a lack of non-traditional underwriting data. \textcolor{black}{Using \ac{NLP} solutions in the insurance industry yields} better customer satisfaction and profit. The use cases include claims classification, optimizing payment processes, monitoring policy changes, personalized product offerings, improved risk assessment, enhanced fraud detection, and business process automation \cite{p2}.

Documents and textual data are rich sources of information that can be used to solve various problems. However, extracting information from this type of data requires more complex techniques due to the unstructured nature of textual data, which can be time-consuming and challenging \cite{p1}.
Text classification is an important step in text processing. This can be done manually by domain experts or automatically. \textcolor{black}{Recently}, due to the increasing number of textual documents, automatic text classification has become popular and important. Automatic text classification approaches are grouped into rule-based and data-driven methods.
Rule-based methods categorize text samples using a set of pre-defined rules that requires deep expert knowledge, which is context-based and hard to acquire. On the other hand, data-driven methods use \ac{ML} approaches to find patterns to classify textual samples \cite{p2}.

However, most of the ML-based text classifiers are built from labeled training \textcolor{black}{samples}. Manual labeling of a large set of training documents is time-consuming. In the past few years, researchers investigated various forms of semi-supervised learning to reduce the burden of manual labeling by using a small labeled set for every class and a large unlabeled set for classifier building. Semi-supervised learning is a hybrid approach that combines supervised and unsupervised learning elements to train models with a small amount of labeled data, and a large amount of unlabeled data \cite{29, 31}. \textcolor{black}{Moreover, word embedding techniques were used to  represent the words of a text using representation learning methods in a way that words that have the same meaning have a similar representation \cite{yin2018dimensionality}.} Similar to word embeddings, distributed representations for sentences can also be learned in an unsupervised fashion by optimizing some auxiliary objectives, such as the reconstruction loss of an autoencoder \cite{31}. Such unsupervised learning results in sentence encoders that can map sentences with similar semantic and syntactic properties to similar fixed-size vector representations. 

In this paper, a novel text classification model, CRL+, that combines \textcolor{black}{\ac{CRL}} and \textcolor{black}{Active Learning} is proposed to handle the challenge of using semi-supervised textual data for text classification in the insurance industry. In this method, supervised \ac{CRL} will be used to train a RoBERTa transformer model to encode the textual data into the representational vector and then classify using a classification layer. This \ac{CRL}-based transformer model will be used as the base model of a modified Active Learning mechanism to classify all the data in an iterative manner.

The remainder of this paper is organized as follows. Section \ref{sec:rw} provides a summary of \textcolor{black}{the  related works in the literature}. Section \ref{sec:bg} provides some background about the \textcolor{black}{methods used in this paper}. Section \ref{sec:pm} describes the proposed approach \textcolor{black}{for classifying} semi-supervised textual data. Section \ref{sec:em} explains the experimental setup, including the dataset and evaluation metrics. Section \ref{sec:res} shows the experimental results. Finally, Section \ref{sec:con} concludes the paper.

\section{Related Works}
\label{sec:rw}
It is estimated that around 80\% of all information is unstructured \cite{DeepTalk}, with text being one of the most common unstructured data types. The unstructured data structure is irregular or incomplete, and there is no predefined data model. Compared to structured data, \textcolor{black}{this data is} still difficult to retrieve, analyze and store \cite{10}. This is where text classification with ML comes in. ML-based techniques can automatically classify all manner of relevant text, from legal documents, social media, surveys, and more, in a fast and cost-effective way \cite{11, 12}. Some of the most popular machine learning algorithms for creating text classification models include the \ac{NB} \cite{mccallum1998comparison} family of algorithms, \ac{SVM} \cite{manevitz2001one}, and \ac{DNN}.  
 
Recent research shows that it is effective to cast many \ac{NLP} tasks as text classification by allowing \ac{DNN} to take a pair of texts as input \cite{13, 14, 15}. Compared to \textcolor{black}{traditional} \ac{ML}, \ac{DNN} algorithms need more training data. However, they \textcolor{black}{do not} have a threshold for learning from training data like traditional machine learning algorithms, such as \ac{SVM} and \ac{NB}. The two main \ac{DNN} architectures for text classification are \ac{CNN} and \ac{RNN}. \ac{RNN}s are trained to recognize patterns across time, whereas \ac{CNN}s learn to recognize patterns across space \cite{16}. \ac{RNN}s work well for the \ac{NLP} tasks such as \ac{QA}, where the comprehension of long-range semantics is required. In contrast, \ac{CNN}s work well where detecting local and position-invariant patterns are essential \cite{p1, 17}. Thus, \ac{RNN}s have become one of \ac{NLP} 's most popular model architectures. \ac{LSTM} is a popular architecture, which addresses the gradient vanishing problems in \ac{RNN}s by introducing a memory cell to remember values over arbitrary time Intervals \cite{18, 19}. There have been works improving \ac{RNN}s and \ac{LSTM} models for \ac{NLP} applications by capturing richer information, such as tree structures of natural language, long-span word relations in text, and document topics \cite{20, 21}. Character-level \ac{CNN}s have also been explored for text classification \cite{22, 23}. Studies investigate the impact of word embeddings and \ac{CNN} architectures on model performance. \cite{24} presented a \ac{VDCNN} model for text processing. It operates directly at the character level and uses only small convolutions and pooling operations. This study shows that the performance of \ac{VDCNN} improves in deeper models. \ac{DNN} algorithms, like word2vec \cite{25} or Glove \cite{26}, are also used to obtain better vector representations for words and improve the accuracy of classifiers trained with traditional machine learning algorithms. \textcolor{black}{Recently, transformer models were introduced to improve the performance of \ac{LSTM} models. Unlike \ac{LSTM}, transformer models can be fully parallelized on GPUs, which makes the training step faster. Moreover, transformers can memorize longer sentences better than the \ac{LSTM}s due to their attention mechanism. AS one of the most popular transformer models, the BERT model was introduced by Google in 2019 \cite{13}. BERT model outperforms its predecessors, Word2Vec and ELMo \cite{28}, exceeding state-of-the-art by a margin in multiple natural language understanding tasks. Section \ref{sec:tran} will discuss the transformers in more detail.}


\section{Background}
\label{sec:bg}
In this chapter, background materials related to this paper will be introduced.

\subsection{Semi-Supervised Learning}
Semi-supervised learning is a type of \ac{ML} in which \textcolor{black}{only a portion of the training samples are labeled.}
Most of the real-world problems in the fintech and insurance industries are semi-supervised. \textcolor{black}{Therefore}, handling this type of problem \textcolor{black}{has recently gained momentum} \cite{p2, p4}. Active Learning and \ac{CRL} are two popular solutions for these problems.

\subsection{Active Learning}
Active Learning is a mechanism to label unlabeled data iteratively using an \ac{ML} model. This mechanism \textcolor{black}{increases the performance of \ac{ML} models on datasets with limited labeled samples}. 
In this mechanism, first, an \ac{ML} model is trained using the labeled data. Then, the model is applied to unlabeled data to classify them. Based on the classification result, some samples are selected to pass to an expert person to label them manually and add them to the labeled data. This process is \textcolor{black}{repeated} until the model meets \textcolor{black}{predefined performance criterion} \cite{p4}.

\subsection{Contrastive Representation Learning}
\ac{CRL} is a \ac{ML} technique used for unsupervised and semi-supervised problems as a pre-training to enhance the performance of \ac{ML} models. This model is used to train a representation space in which similar sample points stay close, while dissimilar ones are far apart. 
\ac{CRL} was first developed as a self-supervised method for unsupervised image classification problems (or as an unsupervised pre-training for supervised problems) \cite{chen2020simple}. In this method, a self-supervised pre-training was done on the data using augmentation techniques to map the samples from the original space to the contrastive space. Then, Khosa et al. \cite{khosla2020supervised} proposed a supervised version of \ac{CRL} and showed its advantages compared to the self-supervised version on image classification problems. In this method, instead of considering the anchor sample and its  augmented one as the positive samples, the samples with similar labels are also considered as positives. 

Moreover, several \ac{NLP} versions of \ac{CRL} were proposed that include both self-supervised and supervised \textcolor{black}{versions} of these models \cite{giorgi2020declutr, gao2021simcse}. In this paper, a supervised \ac{CRL} model, introduced in \cite{sedghamiz2021supcl}, is used as the base model to pre-train the RoBERTa model using the Active Learning mechanism.

\subsection{\ac{RNN}}

\ac{RNN} is a type of neural network that predicts new situations based on previous ones. \ac{RNN}s can handle sequential problems like \ac{NLP}. Several \ac{RNN} models are proposed in the literature. One of the most popular \ac{RNN} models is \ac{LSTM}, \textcolor{black}{which suffers less from the vanishing gradient problem compared to \ac{RNN}} \cite{hochreiter1997long}.

\subsection{Transformers}
\label{sec:tran}

One of the computational bottlenecks \textcolor{black}{of training \ac{RNN}s on GPUs} is the sequential processing of text. Transformers \cite{t5} overcome this limitation by applying self-attention to compute \textcolor{black}{an attention score} in parallel for every word in a sentence or document an "attention score" to model each word's influence on another. Due to this feature, Transformers allow for much more parallelization than \ac{CNN}s and \ac{RNN}s, which makes it possible to efficiently train huge models on large amounts of data on GPUs.

Since 2018, we have seen the rise of a set of large-scale Transformer-based \ac{PLM}s. Compared to earlier contextualized embedding models based on \ac{CNN}s \cite{t143} or \ac{LSTM}s \cite{28}, Transformer-based \ac{PLM}s use much deeper network architectures (e.g., 48-layer Transformers \cite{t144}), and are pre-trained on much larger amounts of text corpora to learn contextual text representations by predicting words conditioned on their context. This \ac{PLM}s are fine-tuned using task-specific labels and have created a new state of the art in many downstream \ac{NLP} tasks, including text classification. Although pre-training is unsupervised (or self-supervised), fine-tuning is supervised learning. A recent survey by Qiu et al. \cite{t145} categorizes popular \ac{PLM}s by their representation types, model architectures, pre-training tasks, and downstream tasks.

\ac{PLM}s can be grouped into two categories, autoregressive and autoencoding \ac{PLM}s. One of the earliest autoregressive \ac{PLM}s is OpenGPT \cite{t6, t144}, a unidirectional model that predicts a text sequence word by word from left to right (or right to left), with each word prediction depending on previous predictions. It consists of 12 layers of Transformer blocks, each consisting of a masked multi-head attention module, followed by a layer normalization and a position-wise feed-forward layer. OpenGPT can be adapted to \ac{NLP} applications such as text classification by adding task-specific linear classifiers and fine-tuning using task-specific labels. 

BERT is one of the most widely used autoencoding \ac{PLM}s \cite{13}. Unlike OpenGPT, which predicts words based on previous predictions, BERT is trained using the \ac{MLM} task that randomly masks some tokens in a text sequence and then independently covers the masked tokens by conditioning on the encoding vectors obtained by a bidirectional Transformer. There have been numerous works on improving BERT. RoBERTa \cite{t146} is more robust than BERT, mainly because its pre-training method focuses on \ac{MLM} with changing mask tokens per epoch, and is trained using much more training data. ALBERT \cite{t147} lowers the memory consumption of the BERT model and increases its training speed. DistillBERT \cite{t148} utilizes knowledge distillation during pre-training to reduce the size of BERT by 40\%, while retaining 99\% of its original capabilities and making the inference 60\% faster. SpanBERT \cite{t149} extends BERT to better represent and predict text spans. Electra \cite{t150} uses a more sample-efficient pre-training task than \ac{MLM}, called replaced token detection. Instead of masking the input, it corrupts it by replacing some tokens with plausible alternatives sampled from a small generator network. ERNIE \cite{t151, t152} incorporates domain knowledge from external knowledge bases, such as named entities, for model pre-training. ALUM \cite{t14} introduces an adversarial loss for model pretraining that improves the model's generalization to new tasks and robustness to adversarial attacks. BERT and its variants have been fine-tuned for various \ac{NLP} tasks, including \ac{QA} \cite{t153}, text classification [154], and \ac{NLI} \cite{15, t155}.

\section{Proposed Method}
\label{sec:pm}
\begin{figure*}
    \centering
    \includegraphics[width=\textwidth]{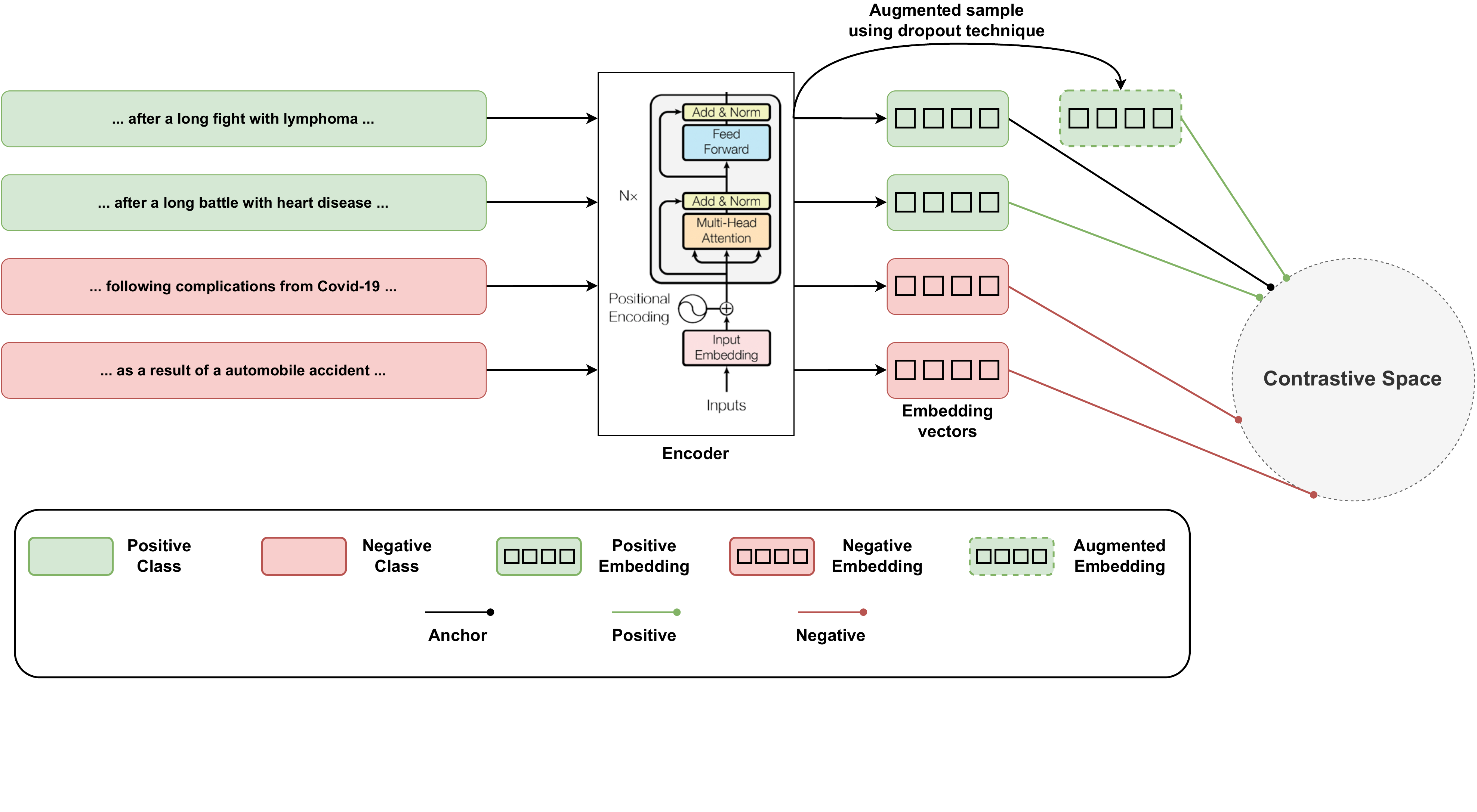}
    \caption{The used Supervised \ac{CRL} model in this work. Samples with similar labels are considered positive samples, and all other samples are considered negative. The augmentation is done using the dropout technique in the encoder.}
    \label{fig:crl}
\end{figure*}

\textcolor{black}{T}his paper \textcolor{black}{combines} supervised \ac{CRL} with an Active Learning mechanism to enhance the performance of these methods to classify textual data. 
In this paper, a modified version of SupCL-Seq \cite{sedghamiz2021supcl} is used as the \ac{CRL} model. SupCL-Seq extends the self-supervised Contrastive Learning for textual data to a supervised setting. In the developed model, several dropouts are used to make augmented samples from the anchor by changing sample embeddings. Then, the anchor sample, its augmented samples, and other samples with the same label in the dataset are used to train the representation using the contrastive loss function. The representation learning task consists of an encoder part of a transformer (RoBERTa) to make the augmented samples and train the \ac{CRL} part of the model. Figure \ref{fig:crl} shows the training of the developed model. Then, a classification layer is added to the trained representation learning model to do the final classification. Finally, all the \ac{CRL} encoder parameters are frozen, and the classification model is trained.

Equation \ref{eq:1} shows the supervised contrastive loss function.
\begin{equation}
\label{eq:1}
    \mathcal{L}^{sup}_i = \sum_{i\in I} \frac{-1}{|p(i)|} \sum_{\textit{p} \in p(i)} log \frac{e^{cosine(\Tilde{x_i},\Tilde{x_p})/\tau}}{\sum_{b\in B(i)} e^{cosine(\Tilde{x_i}, \Tilde{x_b})/\tau}} 
\end{equation}
Where \textcolor{black}{$p(i)$} contains positive samples (samples with a similar label with the anchor), $B(i)$ has negative samples (samples with a different label with the anchor), \textcolor{black}{$\tau$}  is the scaling term, and $cosine(a,b)$ is the similarity function between $a$ and $b$ (see equation \ref{eq:2}).

\begin{equation}
\label{eq:2}
    cosine(A, B) =\frac{A.B}{||A||||B||}=\frac{\sum^{n}_{i=1}A_iB_i}{\sqrt{\sum_{i=1}^{n}A_i^2}\sqrt{\sum_{i=1}^{n}B_i^2}}
\end{equation}

This method is used to pre-train the encoder of the RoBERTa transformer to encode the textual data into a contrastive representation before passing it to the classification layer. 

\begin{figure*}
    \centering
    \includegraphics[width=\textwidth]{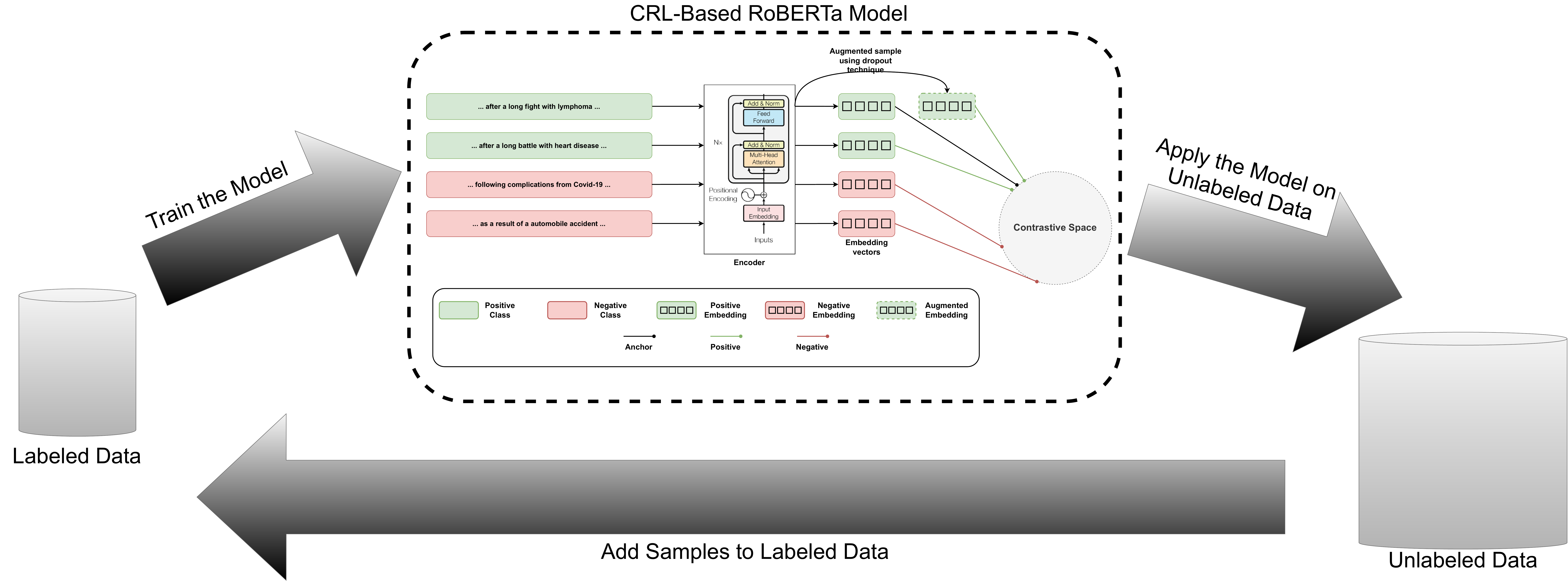}
    \caption{The proposed model consists of the supervised \ac{CRL}-based RoBERTa model as the base model in the Active Learning mechanism. It starts with a small labeled data and tries to label all the samples in several iterations using the \ac{CRL}-based RoBERTa base model.}
    \label{fig:ac}
\end{figure*}

To enhance the performance of this model on semi-supervised applications, this model is used as the base model in the Active Learning mechanism. Figure \ref{fig:ac} shows the proposed model. 

To make the proposed method \textcolor{black}{free from human intervention}, the Active Learning mechanism is changed in a way to remove the expert person from it. In the proposed method, in each iteration of the Active Learning mechanism, the classified samples with reasonable confidence (based on the softmax layer) are added to the labeled samples and used to train the \ac{CRL}-based RoBERTa for the next iteration. This process continues until the model gets a pre-defined performance or a specific number of iterations is passed. 

\section{Experimental Setup}
\label{sec:em}

\subsection{Dataset}
To evaluate the proposed method for labeling and classification of the insurance-based textual data, a dataset consisting of obituary texts is used to predict the cause of death. However, among more than 2,500,000 samples, only 3\% of them were labeled. Table \ref{tbl:labeled_data} shows the number of labeled data for each class in the mentioned dataset.

\begin{table}[]
    \centering
    \caption{Number of the labeled samples for each class label in the obituary dataset.}
    \renewcommand{\arraystretch}{1.5}
    \begin{tabular}{l|c}
         \textbf{Class Label} & \textbf{Number of Labeled Samples}  \\
         \hline
         Neoplasms (Cancers) & 33,104 \\
         Circulatory System & 3,477 \\
         Accidents & 11,942 \\
         Respiratory System & 3,427 \\
         Nervous System & 1,991 \\
         Suicides & 3 \\
         Digestive System & 543 \\
         COVID-19 & 5,007
    \end{tabular}

    \label{tbl:labeled_data}
\end{table}

\subsection{Evaluation Metrics}
The basic \ac{ML} metrics are \ac{TP}, \ac{TN}, \ac{FP}, and \ac{FN}, which represent the number of samples correctly classified as positive, correctly classified as negative, wrongly classified as positive, and wrongly classified as negative, respectively. 
Using these basic metrics, more complex metrics, including Accuracy, Precision, Recall, and F-measure, are defined and used to quantify the performance of \ac{ML} algorithms.
	
	\begin{equation}
		Accuracy=\dfrac{TP+TN}{TP+TN+FP+FN}
		\label{eq:det_acc}
	\end{equation}
	\begin{equation}
		Precision=\dfrac{TP}{TP+FP}
		\label{eq:det_pre}
	\end{equation}
	\begin{equation}
		Recall=\dfrac{TP}{TP+FN}
		\label{eq:det_rec}
	\end{equation}
	\begin{equation}
		F-measure=\dfrac{2\times Precision\times Recall}{Precision+Recall}
		\label{eq:det_fm}
	\end{equation}

	\begin{itemize}
		\item  Accuracy indicates the number of correctly classified samples over the entire dataset (see equation \ref{eq:det_acc}).
		
		\item Precision indicates the number of samples classified correctly as each class label over total samples classified for that class (see equation \ref{eq:det_pre}).
		
		\item Recall indicates the number of samples classified  as each class label correctly over the total instances of the dataset for that class label  (see equation \ref{eq:det_rec}).
		
		\item F-measure is the harmonic value of precision and recall (see equation \ref{eq:det_fm}).

	\end{itemize}


\section{Results}
\label{sec:res}

The proposed combination of the supervised \ac{CRL}, RoBERTa transformer, and Active Learning is evaluated on an obituary dataset introduced in Section \ref{sec:em}. Table \ref{tbl:pmres} shows the performance of the proposed method. This table illustrates that the proposed method can accurately classify textual insurance data for all class labels.
Moreover, the proposed method is compared with the RoBERTa transformer with supervised \ac{CRL} and Active Learning with the RoBERTa base model. As illustrated in Table \ref{tbl:compare}, the proposed method outperformed both these models. This table shows that the supervised \ac{CRL} model significantly outperformed the Active Learning model. However, adding Active Learning to this model empowers it to detect the cause of death more accurately.

\begin{table}[]
\centering
\caption{Results of the proposed method on the obituary dataset. The overall row shows the performance of the model over all the classes. The other rows show the one-versus-all version of the proposed method for each class label.}
\renewcommand{\arraystretch}{1.5}
\begin{adjustbox}{width=0.45\textwidth} 
\begin{tabular}{l|c|c|c|c}

\textbf{Class Label} & \textbf{Accuracy} & \textbf{Precision} & \textbf{Recall} & \textbf{f-measure} \\ \hline
\textbf{Overall}     & \textbf{95.68}             & \textbf{95.25}              & \textbf{95.68}           &   \textbf{95.46}                 \\ \hline
Neoplasms (Cancers)  & 96.72             & 98.50              & 96.72           &   97.60                 \\ \hline
Covid-19             & 99.00             & 92.25              & 96.30           & 94.23              \\ \hline
Circulatory System   & 99.11             & 93.28              & 91.20           & 92.23              \\ \hline
Accidents            & 98.26             & 96.35              & 95.00           & 95.67              \\ \hline
Respiratory System   & 96.17             & 59.44              & 87.70           & 66.95              \\ \hline
Nervous System       & 99.70             & 95.59              & 95.20           & 95.57              \\ \hline
Suicides             & 100               & 100                & 100             & 100                \\ \hline
Digestive System     & 99.20             & 58.65              & 71.60           & 64.50      
\end{tabular}
\end{adjustbox}
\label{tbl:pmres}
\end{table}

\begin{table}[]
\centering
\caption{Comparing the performance of the proposed method (CRL+) with \ac{CRL} and Active Learning models.}
\renewcommand{\arraystretch}{1.5}
\begin{adjustbox}{width=0.45\textwidth} 
\begin{tabular}{l|c|c|c|c}

\textbf{Model}           & \textbf{Accuracy} & \textbf{Precision} & \textbf{Recall} & \textbf{f-measure} \\ \hline
\textbf{CRL+} & \textbf{95.68}    & \textbf{95.25}     & \textbf{95.68}  & \textbf{95.46}     \\ \hline
\ac{CRL}             & 92.77    & 92.62     & 92.77  & 92.69     \\ \hline
Active Learning & 75.28    & 90.78     & 75.28  & 82.31     \\ 
\end{tabular}
\end{adjustbox}
\label{tbl:compare}
\end{table}

\section{Conclusion}
\label{sec:con}
Insurance companies gathered enormous amounts of textual data through different channels. This information can help insurance companies to \textcolor{black}{perform highly complex advanced analytics} using data mining and machine learning. In this \textcolor{black}{paper}, CRL+, a semi-supervised active contrastive representation learning model is proposed to map the semi-supervised data into a contrastive space. This learned representation can be used for classification or regression models and also sequence-to-sequence applications. The proposed method is evaluated using an obituary dataset to classify the cause of death based on the document's content and compared with a modifier Active Learning and contrastive representation learning methods using RoBERTa base models. The experiments show that the proposed method outperforms the others in all metrics.
The proposed method could be improved by using a two-step pre-processing and adding self-supervised contrastive learning. Moreover, the Active Learning part of the algorithm could be modified to make it more efficient.
	
	\ifCLASSOPTIONcaptionsoff
	\newpage
	\fi

	\bibliographystyle{IEEEtran}
	\bibliography{Ref}

\begin{thebibliography}{10}
\providecommand{\url}[1]{#1}
\csname url@samestyle\endcsname
\providecommand{\newblock}{\relax}
\providecommand{\bibinfo}[2]{#2}
\providecommand{\BIBentrySTDinterwordspacing}{\spaceskip=0pt\relax}
\providecommand{\BIBentryALTinterwordstretchfactor}{4}
\providecommand{\BIBentryALTinterwordspacing}{\spaceskip=\fontdimen2\font plus
\BIBentryALTinterwordstretchfactor\fontdimen3\font minus
  \fontdimen4\font\relax}
\providecommand{\BIBforeignlanguage}[2]{{%
\expandafter\ifx\csname l@#1\endcsname\relax
\typeout{** WARNING: IEEEtran.bst: No hyphenation pattern has been}%
\typeout{** loaded for the language `#1'. Using the pattern for}%
\typeout{** the default language instead.}%
\else
\language=\csname l@#1\endcsname
\fi
#2}}
\providecommand{\BIBdecl}{\relax}
\BIBdecl

\bibitem{p1}
S.~Minaee, N.~Kalchbrenner, E.~Cambria, N.~Nikzad, M.~Chenaghlu, and J.~Gao,
  ``Deep learning--based text classification: a comprehensive review,''
  \emph{ACM Computing Surveys (CSUR)}, vol.~54, no.~3, pp. 1--40, 2021.

\bibitem{p2}
A.~Ly, B.~Uthayasooriyar, and T.~Wang, ``A survey on natural language
  processing (nlp) and applications in insurance,'' \emph{arXiv preprint
  arXiv:2010.00462}, 2020.

\bibitem{29}
A.~Rasmus, M.~Berglund, M.~Honkala, H.~Valpola, and T.~Raiko, ``Semi-supervised
  learning with ladder networks,'' \emph{Advances in neural information
  processing systems}, vol.~28, 2015.

\bibitem{31}
D.~E. Rumelhart, G.~E. Hinton, and R.~J. Williams, ``Learning internal
  representations by error propagation,'' California Univ San Diego La Jolla
  Inst for Cognitive Science, Tech. Rep., 1985.

\bibitem{yin2018dimensionality}
Z.~Yin and Y.~Shen, ``On the dimensionality of word embedding,'' \emph{Advances
  in neural information processing systems}, vol.~31, 2018.

\bibitem{DeepTalk}
\BIBentryALTinterwordspacing
{Deep Talk}, ``80\% of the world's data is unstructured.'' [Online]. Available:
  \url{https://deep-talk.medium.com/80-of-the-worlds-data-is-unstructured-7278e2ba6b73}
\BIBentrySTDinterwordspacing

\bibitem{10}
S.~Gupta and S.~Gupta, ``Natural language processing in mining unstructured
  data from software repositories: a review,'' \emph{S{\=a}dhan{\=a}}, vol.~44,
  no.~12, pp. 1--17, 2019.

\bibitem{11}
M.~P. Akhter, Z.~Jiangbin, I.~R. Naqvi, M.~Abdelmajeed, A.~Mehmood, and M.~T.
  Sadiq, ``Document-level text classification using single-layer multisize
  filters convolutional neural network,'' \emph{IEEE Access}, vol.~8, pp.
  42\,689--42\,707, 2020.

\bibitem{12}
H.~Liu, P.~Burnap, W.~Alorainy, and M.~L. Williams, ``A fuzzy approach to text
  classification with two-stage training for ambiguous instances,'' \emph{IEEE
  Transactions on Computational Social Systems}, vol.~6, no.~2, pp. 227--240,
  2019.

\bibitem{mccallum1998comparison}
A.~McCallum, K.~Nigam \emph{et~al.}, ``A comparison of event models for naive
  bayes text classification,'' in \emph{AAAI-98 workshop on learning for text
  categorization}, vol. 752, no.~1.\hskip 1em plus 0.5em minus 0.4em\relax
  Madison, WI, 1998, pp. 41--48.

\bibitem{manevitz2001one}
L.~M. Manevitz and M.~Yousef, ``One-class svms for document classification,''
  \emph{Journal of machine Learning research}, vol.~2, no. Dec, pp. 139--154,
  2001.

\bibitem{13}
J.~Devlin, M.-W. Chang, K.~Lee, and K.~Toutanova, ``Bert: Pre-training of deep
  bidirectional transformers for language understanding,'' \emph{arXiv preprint
  arXiv:1810.04805}, 2018.

\bibitem{14}
A.~Wang, A.~Singh, J.~Michael, F.~Hill, O.~Levy, and S.~R. Bowman, ``Glue: A
  multi-task benchmark and analysis platform for natural language
  understanding,'' \emph{arXiv preprint arXiv:1804.07461}, 2018.

\bibitem{15}
X.~Liu, P.~He, W.~Chen, and J.~Gao, ``Multi-task deep neural networks for
  natural language understanding,'' \emph{arXiv preprint arXiv:1901.11504},
  2019.

\bibitem{16}
Y.~LeCun, L.~Bottou, Y.~Bengio, and P.~Haffner, ``Gradient-based learning
  applied to document recognition,'' \emph{Proceedings of the IEEE}, vol.~86,
  no.~11, pp. 2278--2324, 1998.

\bibitem{17}
N.~Kalchbrenner, E.~Grefenstette, and P.~Blunsom, ``A convolutional neural
  network for modelling sentences,'' \emph{arXiv preprint arXiv:1404.2188},
  2014.

\bibitem{18}
Y.~Liu, C.~Sun, L.~Lin, and X.~Wang, ``Learning natural language inference
  using bidirectional lstm model and inner-attention,'' \emph{arXiv preprint
  arXiv:1605.09090}, 2016.

\bibitem{19}
P.~Liu, X.~Qiu, and X.~Huang, ``Modelling interaction of sentence pair with
  coupled-lstms,'' \emph{arXiv preprint arXiv:1605.05573}, 2016.

\bibitem{20}
K.~S. Tai, R.~Socher, and C.~D. Manning, ``Improved semantic representations
  from tree-structured long short-term memory networks,'' \emph{arXiv preprint
  arXiv:1503.00075}, 2015.

\bibitem{21}
X.~Zhu, P.~Sobihani, and H.~Guo, ``Long short-term memory over recursive
  structures,'' in \emph{International Conference on Machine Learning}.\hskip
  1em plus 0.5em minus 0.4em\relax PMLR, 2015, pp. 1604--1612.

\bibitem{22}
X.~Zhang, J.~Zhao, and Y.~LeCun, ``Character-level convolutional networks for
  text classification,'' \emph{Advances in neural information processing
  systems}, vol.~28, 2015.

\bibitem{23}
Y.~Kim, Y.~Jernite, D.~Sontag, and A.~M. Rush, ``Character-aware neural
  language models,'' in \emph{Thirtieth AAAI conference on artificial
  intelligence}, 2016.

\bibitem{24}
A.~Conneau, H.~Schwenk, L.~Barrault, and Y.~Lecun, ``Very deep convolutional
  networks for text classification,'' \emph{arXiv preprint arXiv:1606.01781},
  2016.

\bibitem{25}
T.~Mikolov, K.~Chen, G.~Corrado, and J.~Dean, ``Efficient estimation of word
  representations in vector space,'' \emph{arXiv preprint arXiv:1301.3781},
  2013.

\bibitem{26}
J.~Pennington, R.~Socher, and C.~D. Manning, ``Glove: Global vectors for word
  representation,'' in \emph{Proceedings of the 2014 conference on empirical
  methods in natural language processing (EMNLP)}, 2014, pp. 1532--1543.

\bibitem{28}
M.~E. Peters, M.~Neumann, M.~Iyyer, M.~Gardner, C.~Clark, K.~Lee, and
  L.~Zettlemoyer, ``Deep contextualized word representations,'' in
  \emph{Proceedings of the 2018 Conference of the North {A}merican Chapter of
  the Association for Computational Linguistics: Human Language Technologies,
  Volume 1 (Long Papers)}.\hskip 1em plus 0.5em minus 0.4em\relax New Orleans,
  Louisiana: Association for Computational Linguistics, Jun. 2018, pp.
  2227--2237.

\bibitem{p4}
R.~Elie, C.~Hillairet, F.~Hu, and M.~Juillard, ``An overview of active learning
  methods for insurance with fairness appreciation,'' \emph{arXiv preprint
  arXiv:2112.09466}, 2021.

\bibitem{chen2020simple}
T.~Chen, S.~Kornblith, M.~Norouzi, and G.~Hinton, ``A simple framework for
  contrastive learning of visual representations,'' in \emph{International
  conference on machine learning}.\hskip 1em plus 0.5em minus 0.4em\relax PMLR,
  2020, pp. 1597--1607.

\bibitem{khosla2020supervised}
P.~Khosla, P.~Teterwak, C.~Wang, A.~Sarna, Y.~Tian, P.~Isola, A.~Maschinot,
  C.~Liu, and D.~Krishnan, ``Supervised contrastive learning,'' \emph{Advances
  in Neural Information Processing Systems}, vol.~33, pp. 18\,661--18\,673,
  2020.

\bibitem{giorgi2020declutr}
J.~Giorgi, O.~Nitski, B.~Wang, and G.~Bader, ``Declutr: Deep contrastive
  learning for unsupervised textual representations,'' \emph{arXiv preprint
  arXiv:2006.03659}, 2020.

\bibitem{gao2021simcse}
T.~Gao, X.~Yao, and D.~Chen, ``Simcse: Simple contrastive learning of sentence
  embeddings,'' \emph{arXiv preprint arXiv:2104.08821}, 2021.

\bibitem{sedghamiz2021supcl}
H.~Sedghamiz, S.~Raval, E.~Santus, T.~Alhanai, and M.~Ghassemi, ``Supcl-seq:
  Supervised contrastive learning for downstream optimized sequence
  representations,'' \emph{arXiv preprint arXiv:2109.07424}, 2021.

\bibitem{hochreiter1997long}
S.~Hochreiter and J.~Schmidhuber, ``Long short-term memory,'' \emph{Neural
  computation}, vol.~9, no.~8, pp. 1735--1780, 1997.

\bibitem{t5}
A.~Vaswani, N.~Shazeer, N.~Parmar, J.~Uszkoreit, L.~Jones, A.~N. Gomez,
  {\L}.~Kaiser, and I.~Polosukhin, ``Attention is all you need,''
  \emph{Advances in neural information processing systems}, vol.~30, 2017.

\bibitem{t143}
R.~Collobert, J.~Weston, L.~Bottou, M.~Karlen, K.~Kavukcuoglu, and P.~Kuksa,
  ``Natural language processing (almost) from scratch,'' \emph{Journal of
  machine learning research}, vol.~12, no. ARTICLE, pp. 2493--2537, 2011.

\bibitem{t144}
A.~Radford, J.~Wu, R.~Child, D.~Luan, D.~Amodei, I.~Sutskever \emph{et~al.},
  ``Language models are unsupervised multitask learners,'' \emph{OpenAI blog},
  vol.~1, no.~8, p.~9, 2019.

\bibitem{t145}
X.~Qiu, T.~Sun, Y.~Xu, Y.~Shao, N.~Dai, and X.~Huang, ``Pre-trained models for
  natural language processing: A survey,'' \emph{Science China Technological
  Sciences}, vol.~63, no.~10, pp. 1872--1897, 2020.

\bibitem{t6}
A.~Radford, K.~Narasimhan, T.~Salimans, I.~Sutskever \emph{et~al.}, ``Improving
  language understanding by generative pre-training,'' 2018.

\bibitem{t146}
Y.~Liu, M.~Ott, N.~Goyal, J.~Du, M.~Joshi, D.~Chen, O.~Levy, M.~Lewis,
  L.~Zettlemoyer, and V.~Stoyanov, ``Roberta: A robustly optimized bert
  pretraining approach,'' \emph{arXiv preprint arXiv:1907.11692}, 2019.

\bibitem{t147}
Z.~Lan, M.~Chen, S.~Goodman, K.~Gimpel, P.~Sharma, and R.~Soricut, ``Albert: A
  lite bert for self-supervised learning of language representations,''
  \emph{arXiv preprint arXiv:1909.11942}, 2019.

\bibitem{t148}
V.~Sanh, L.~Debut, J.~Chaumond, and T.~Wolf, ``Distilbert, a distilled version
  of bert: smaller, faster, cheaper and lighter,'' \emph{arXiv preprint
  arXiv:1910.01108}, 2019.

\bibitem{t149}
M.~Joshi, D.~Chen, Y.~Liu, D.~S. Weld, L.~Zettlemoyer, and O.~Levy, ``Spanbert:
  Improving pre-training by representing and predicting spans,''
  \emph{Transactions of the Association for Computational Linguistics}, vol.~8,
  pp. 64--77, 2020.

\bibitem{t150}
K.~Clark, M.-T. Luong, Q.~V. Le, and C.~D. Manning, ``Electra: Pre-training
  text encoders as discriminators rather than generators,'' \emph{arXiv
  preprint arXiv:2003.10555}, 2020.

\bibitem{t151}
Y.~Sun, S.~Wang, Y.~Li, S.~Feng, X.~Chen, H.~Zhang, X.~Tian, D.~Zhu, H.~Tian,
  and H.~Wu, ``Ernie: Enhanced representation through knowledge integration,''
  \emph{arXiv preprint arXiv:1904.09223}, 2019.

\bibitem{t152}
Y.~Sun, S.~Wang, Y.~Li, S.~Feng, H.~Tian, H.~Wu, and H.~Wang, ``Ernie 2.0: A
  continual pre-training framework for language understanding,'' in
  \emph{Proceedings of the AAAI Conference on Artificial Intelligence},
  vol.~34, no.~05, 2020, pp. 8968--8975.

\bibitem{t14}
X.~Liu, H.~Cheng, P.~He, W.~Chen, Y.~Wang, H.~Poon, and J.~Gao, ``Adversarial
  training for large neural language models,'' \emph{arXiv preprint
  arXiv:2004.08994}, 2020.

\bibitem{t153}
S.~Garg, T.~Vu, and A.~Moschitti, ``Tanda: Transfer and adapt pre-trained
  transformer models for answer sentence selection,'' in \emph{Proceedings of
  the AAAI Conference on Artificial Intelligence}, vol.~34, no.~05, 2020, pp.
  7780--7788.

\bibitem{t155}
Z.~Zhang, Y.~Wu, H.~Zhao, Z.~Li, S.~Zhang, X.~Zhou, and X.~Zhou,
  ``Semantics-aware bert for language understanding,'' in \emph{Proceedings of
  the AAAI Conference on Artificial Intelligence}, vol.~34, no.~05, 2020, pp.
  9628--9635.

\end{thebibliography}

\end{document}